\documentclass[10pt,twocolumn,letterpaper]{article}

\usepackage{cvpr}
\usepackage{times}
\usepackage{epsfig}
\usepackage{graphicx}
\usepackage{amsmath}
\usepackage{amssymb}

\usepackage{pifont}
\newcommand{\cmark}{\ding{51}}%
\newcommand{\xmark}{\ding{55}}%

\usepackage{color}
\usepackage{multirow}
\usepackage{caption}
\usepackage{subcaption}
\usepackage{tabularx} 
\usepackage{algorithm}  
\usepackage{algorithmic} 

\usepackage{amsthm}

\theoremstyle{definition}

\theoremstyle{remark}

\usepackage{algorithm}  
\usepackage{algorithmic} 

\usepackage{flushend} 

\usepackage[pagebackref=true,breaklinks=true,letterpaper=true,colorlinks,bookmarks=false]{hyperref}

\cvprfinalcopy 

\def\cvprPaperID{216} 

\ifcvprfinal\fi
\begin{document}

\title{Understanding Humans in Crowded Scenes: Deep Nested Adversarial Learning and A New Benchmark for Multi-Human Parsing}

\author{\normalsize{Jian~Zhao$^{1,2}$\thanks{indicates equal contributions.} \quad Jianshu~Li$^{1*}$ \quad Yu~Cheng$^{1*}$ \quad Li~Zhou$^{1}$ \quad Terence~Sim$^{1}$ \quad Shuicheng~Yan$^{1,3}$ \quad Jiashi~Feng$^{1}$}\\
	\small{$^{1}$National University of Singapore \quad $^{2}$National University of Defense Technology \quad  $^{3}$Qihoo 360 AI Institute} \\
	{\small  \{zhaojian90, 	jianshu\}@u.nus.edu \quad chengyu996@gmail.com  \quad zhouli2025@gmail.com} \\ {\small tsim@comp.nus.edu.sg  \quad \{eleyans, elefjia\}@nus.edu.sg}}

\maketitle

\begin{abstract}
	Despite the noticeable progress in perceptual tasks like detection, instance segmentation and human parsing, computers still perform unsatisfactorily on visually understanding humans in crowded scenes, such as group behavior analysis, person re-identification and autonomous driving, \emph{etc}. To this end, models need to comprehensively perceive the semantic information and the differences between instances in a multi-human image, which is recently defined as the \emph{multi-human parsing} task.  In this paper, we present a new large-scale database ``\textbf{M}ulti-\textbf{H}uman \textbf{P}arsing (MHP)" for algorithm development and evaluation, and advances the state-of-the-art in understanding humans in crowded scenes. MHP contains 25{,}403 elaborately annotated images with 58 fine-grained semantic category labels,  involving 2-26 persons per image and captured in real-world scenes from various viewpoints, poses, occlusion, interactions and background. We further propose a novel deep \textbf{N}ested \textbf{A}dversarial \textbf{N}etwork (NAN) model for multi-human parsing. NAN consists of three \textbf{G}enerative \textbf{A}dversarial \textbf{N}etwork (GAN)-like sub-nets, respectively performing semantic saliency prediction, instance-agnostic parsing and instance-aware clustering. These sub-nets form a nested structure and are carefully designed to learn jointly in an end-to-end way. NAN consistently outperforms existing state-of-the-art solutions on our MHP and several other datasets, and serves as a strong baseline to drive the future research for multi-human parsing.
\end{abstract}

\section{Introduction}

\begin{figure}[t]
	\begin{center}
		\includegraphics[width=1\linewidth]{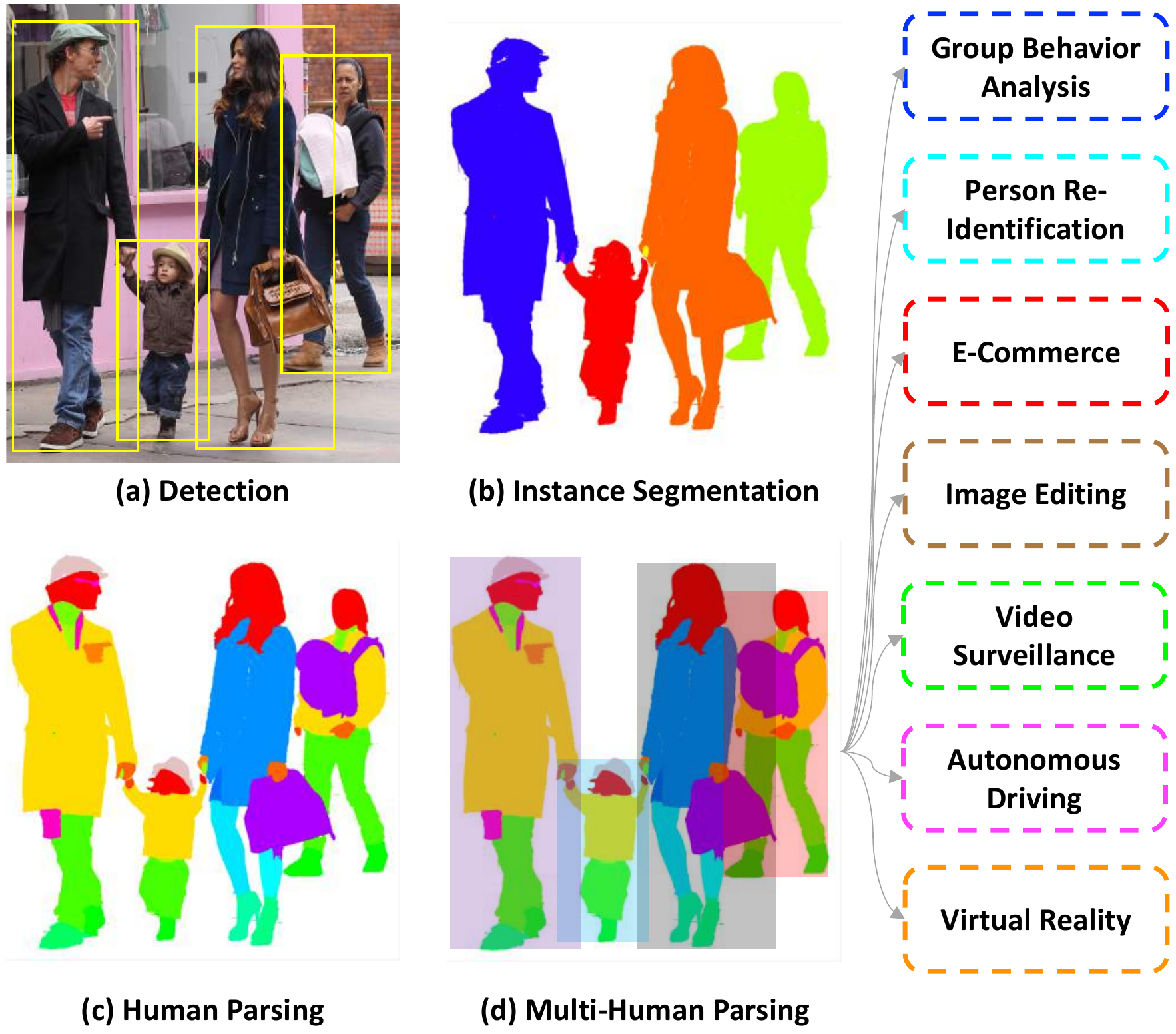}
	\end{center}
	\vspace{-5mm}
	\small
	\caption{\small Illustration of motivation. While existing efforts on human-centric analysis have been devoted to (a) detection (localizing different persons at a coarse, bounding box level), (b) instance segmentation (labelling each pixel of each person uniquely) or (c) human parsing (decomposing persons into their semantic categories), we focus on (d) multi-human parsing (parsing body parts and fashion items at the instance level), which aligns better with many real-world applications. We introduce a new lage-scale, richly-annotated \textbf{M}ulti-\textbf{H}uman \textbf{P}arsing (MHP) dataset consisting of images with various viewpoints, poses, occlusion, human interactions and background. We further propose a novel deep \textbf{N}ested \textbf{A}dversarial \textbf{N}etwork (NAN) model for solving the challenging multi-human parsing problem effectively and efficiently. Best viewed in color.}
	\label{fig: Fig1}
	\vspace{-2mm}
\end{figure}

\begin{table*}[t]
	\newcommand{\tabincell}[2]{\begin{tabular}{@{}#1@{}}#2\end{tabular}}
	\begin{center}
		\scriptsize
		\caption{\small Statistics for publicly available human parsing datasets.}
		\label{tab: Tab1}
		\vspace{-3mm}
		\begin{tabular}{ccccccc}
			\hline
			{Datasets} & {Instance Aware?} & {\# Total} & {\# Training} & {\# Validation} & {\# Testing} & {\# Category} \\
			\hline
			Buffy~\cite{vineet2011human} & \cmark & 748 & 452 & - & 296 & 13 \\
			Fashionista~\cite{yamaguchi2012parsing} & \xmark & 685 & 456 & - & 229 & 56 \\
			PASCAL-Person-Part~\cite{chen2014detect} & \xmark & 3{,}533 & 1{,}716 & - & 1{,}817 & 7 \\
			ATR~\cite{liang2015human} & \xmark & 17{,}700 & 16{,}000 & 700 & 1{,}000 & 18 \\
			LIP~\cite{gong2017look} & \xmark & 50{,}462 & 30{,}462 & 10{,}000 & 10{,}000 & 20 \\
			MHP v1.0~\cite{li2017towards} & \cmark & 4{,}980 & 3{,}000 & 1{,}000 & 980 & 19 \\
			\hline
			MHP v2.0 & \cmark & 25{,}403 & 15{,}403 & 5{,}000 & 5{,}000 & 59 \\			
			\hline
		\end{tabular}
	\end{center}
	\vspace{-6mm}
\end{table*}

One of the primary goals of intelligent human-computer interaction is understanding the humans in visual scenes. It involves several perceptual tasks including detection, \emph{i.e.} localizing different persons at a coarse, bounding box level (Fig.~\ref{fig: Fig1} (a)), instance segmentation, \emph{i.e.} labelling each pixel of each person uniquely (Fig.~\ref{fig: Fig1} (b)), and human parsing, \emph{i.e.}  decomposing persons into their semantic categories (Fig.~\ref{fig: Fig1} (c)). Recently, deep learning based methods have achieved remarkable sucess in these perceptual tasks thanks to the availability of plentiful annotated images for training and evaluation purposes~\cite{dollar2012pedestrian, everingham2015pascal, lin2014microsoft, gong2017look}.

Though exciting, current progress is still far from the utimate goal of visually understanding humans. As Fig.~\ref{fig: Fig1} shows, previous efforts on understanding humans in visual scenes either only consider coarse information or are agnostic to different instances.  In the real-world scenarios, it is more likely that there simutaneously exist multiple persons, with various human interactions, poses and occlusion. Thus, it is more practically demanded to parse human body parts and fashion items at the instance level, which is recently defined as the \emph{multi-human parsing} task~\cite{li2017towards}. Multi-human parsing enables more detailed understanding of humans in crowded scenes and aligns better with many real-world applications, such as group behavior analysis~\cite{gan2016concepts}, person re-identification~\cite{zhao2013unsupervised}, e-commerce~\cite{turban2002electronic}, image editing~\cite{xu2016deep}, video surveillance~\cite{collins2000system}, autonomous driving~\cite{cordts2016cityscapes} and virtual reality~\cite{lin2016virtual}. However, the existing benchmark datasets~\cite{dollar2012pedestrian, everingham2015pascal, lin2014microsoft, gong2017look} are not suitable for such a new task. Even though Li \emph{et al.}~\cite{li2017towards} proposed a preliminary \textbf{M}ulti-\textbf{H}uman \textbf{P}arsing (MHP v1.0) dataset, it only contains 4{,}980 images annotated with 18 semantic labels. In this work, we propose a new large-scale benchmark ``\textbf{M}ulti-\textbf{H}uman \textbf{P}arsing (MHP v2.0)", aiming to push the frontiers of multi-human parsing research towards holistically understanding humans in crowded scenes. The data in MHP v2.0 cover wide variability and complexity \emph{w.r.t.} viewpoints, poses, occlusion, human interactions and background. It in total includes 25{,}403 human images with pixel-wise annotations of 58 semantic categories.

We further propose a novel deep \textbf{N}ested \textbf{A}dversarial \textbf{N}etwork (NAN) model for solving the challenging multi-human parsing problem. Unlike most existing methods~\cite{li2017towards, jiang2016detangling, li2017holistic} which rely on separate stages of instance localization, human parsing and result refinement, the proposed NAN parses semantic categories and differentiates different person instances simultaneously in an effective and time-efficient manner. NAN consists of three \textbf{G}enerative \textbf{A}dversarial \textbf{N}etwork (GAN)-like sub-nets, respectively performing semantic saliency prediction, instance-agnostic parsing and instance-aware clustering. Each sub-task is simpler than the original multi-human parsing task, and is more easily addressed by the corresponding sub-net. Unlike many multi-task learning applications, in our method the sub-nets depend on each other, forming a causal nest by dynamically boosting each other through an adversarial strategy (See Fig.~\ref{fig: Fig4}), which is hence called a ``nested adversarial learning" structure. Such a structure enables effortless gradient \textbf{B}ackpro\textbf{P}agation (BP) in NAN such that it can be trained in a holistic, end-to-end way, which is favorable to both accuracy and speed. We conduct qualitative and quantitative experiments on the MHP v2.0 dataset proposed in this work, as well as the MHP v1.0~\cite{li2017towards}, PASCAL-Person-Part~\cite{chen2014detect} and Buffy~\cite{vineet2011human} benchmark datasets. The results demonstrate the superiority of NAN on multi-human parsing over the state-of-the-arts.

Our contributions are summarized as follows\footnote{The dataset, annotation tools, and source codes for NAN and evaluation metrics are available at \url{https://github.com/ZhaoJ9014/Multi-Human-Parsing_MHP}}.
\begin{itemize}
	\setlength\itemsep{0em}
	\item We propose a new large-scale benchmark and evaluation server to advance understanding of humans in crowded scenes, which contains 25{,}403 images annotated pixel-wisely with 58 semantic category labels.
	\item We propose a novel deep \textbf{N}ested \textbf{A}dversarial \textbf{N}etwork (NAN) model for multi-human parsing, which serves as a strong baseline to inspire more future research efforts on this task.
	\item Comprehensive evaluations on the MHP v2.0 dataset proposed in this work, as well as the MHP v1.0~\cite{li2017towards}, PASCAL-Person-Part~\cite{chen2014detect} and Buffy~\cite{vineet2011human} benchmark datasets verify the superiority of NAN on understanding humans in crowded scenes over the state-of-the-arts.
\end{itemize}

\begin{figure*}[t]
	\begin{center}
		\includegraphics[width=1\linewidth]{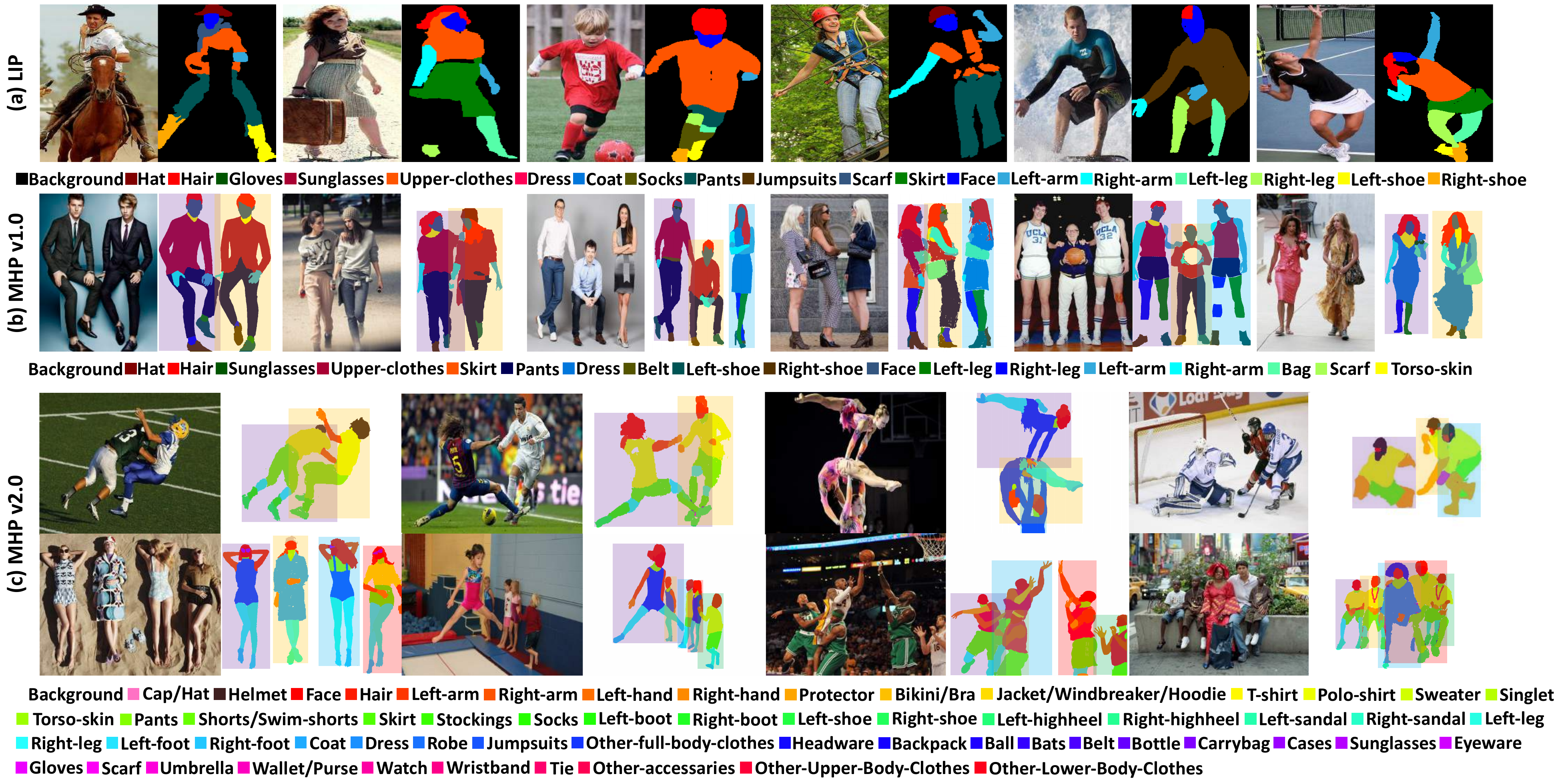}
	\end{center}
	\vspace{-5mm}
	\small
	\caption{\small Annotation examples for our ``\textbf{M}ulti-\textbf{H}uman \textbf{P}arsing (MHP v2.0)" dataset and existing datasets. (a) Examples in LIP~\cite{gong2017look}. LIP is restricted to an instance-agnostic setting and  has limited semantic category annotations. (b) Examples in MHP v1.0~\cite{li2017towards}. MHP v1.0 has lower scalability, variability and complexity, and only contains coarse labels. (c) Examples in our MHP v2.0. MHP v2.0 contains fine-grained semantic category labels with various viewpoints, poses, occlusion, interactions and background, aligned better with reality. Best viewed in color.}
	\label{fig: Fig2}
	\vspace{-3mm}
\end{figure*}

\section{Related Work}

\paragraph{Human Parsing Datasets} The statistics of popular publicly available datasets for human parsing are summarized in Tab.~\ref{tab: Tab1}. The Buffy~\cite{vineet2011human} dataset was released in 2011 for human parsing and instance segmentation. It  contains only 748 images annotated with 13 semantic categories. The Fashionista~\cite{yamaguchi2012parsing} dataset was released in 2012 for human parsing, containing limited images annotated with 56 fashion categories. The PASCAL-Person-Part~\cite{chen2014detect} dataset was initially annotated by Chen \emph{et al.}~\cite{chen2014detect} from the PASCAL-VOC-2010~\cite{everingham2010pascal} dataset. Chen \emph{et al.}~\cite{chen2016attention} extended it for human parsing with 7 coarse body part labels. The ATR~\cite{liang2015human} dataset was released in 2015 for human parsing with a large number of images annotated with 18 semantic categories. The LIP~\cite{gong2017look} dataset further extended ATR~\cite{liang2015human} by cropping person instances from Microsoft COCO~\cite{lin2014microsoft}. It is a large-scale human parsing dataset with densely pixel-wise annotations of 20 semantic categories. But it has two limitations.~1) Despite the large data size, it contains limited semantic category annotations, which restricts the fine-grained understanding of humans in visual scenes. ~2) In LIP~\cite{gong2017look}, only a small proportion of images involve multiple persons with interactions. Such an instance-agnostic setting severely deviates from reality. Even in the MHP v1.0 dataset proposed by Li \emph{et al.}~\cite{li2017towards} for multi-human parsing, only 4{,}980 images are included and annotated with 18 semantic labels. Comparatively, our MHP v2.0 dataset contains 25{,}403 elaborately annotated images with 58 fine-grained semantic part labels. It  is the largest and most comprehensive multi-human parsing dataset to date, to our best knowledge. Visual comparisons between LIP~\cite{gong2017look}, MHP v1.0~\cite{li2017towards} and our MHP v2.0 are provided in Fig.~\ref{fig: Fig2}.

\vspace{-2mm}
\paragraph{Human Parsing Approaches} Recently, many research efforts have been devoted to human parsing~\cite{liang2015proposal, gong2017look, liu2017surveillance, zhao2017self, he2017mask, de2017semantic, li2017towards, jiang2016detangling, li2017holistic} due to its wide range of potential applications. For example, Liang \emph{et al.}~\cite{liang2015proposal} proposed a proposal-free network
for instance segmentation by directly predicting the instance numbers of different categories and the pixel-level information. Gong \emph{et al.}~\cite{gong2017look} proposed a self-supervised structure-sensitive learning approach, which imposes human pose structures to parsing
results without resorting to extra supervision. Liu \emph{et al.}~\cite{liu2017surveillance} proposed a single frame video parsing method which integrates frame parsing, optical flow estimation and temporal fusion into a unified network. Zhao \emph{et al.}~\cite{zhao2017self} proposed a self-supervised neural aggregation
network, which learns to aggregate the multi-scale features and incorporates a self-supervised joint loss to ensure the consistency between parsing and pose. He \emph{et al.}~\cite{he2017mask} proposed the Mask R-CNN, which is extended from Faster R-CNN~\cite{ren2015faster} by adding a branch for predicting an object mask in parallel with the existing branch for bounding box recognition. Brabandere \emph{et al.}~\cite{de2017semantic} proposed to tackle instance segmentation with a discriminative loss function, operating at the pixel level, which encourages a convolutional network to produce a representation of the image that can be easily clustered into instances with a simple post-processing step. However, these methods either only consider coarse semantic information or are agnostic to different instances. To enable more detailed human-centric analysis, Li \emph{et al.}~\cite{li2017towards} initially proposed the multi-human parsing task, which aligns better with the realistic scenarios. They also proposed a novel MH-Parser model as a reference method which generates parsing maps and instance masks simutaneously in a bottom-up fashion. Jiang \emph{et al.}~\cite{jiang2016detangling} proposed a new approach to segment human instances and label their body parts using region assembly. Li \emph{et al.}~\cite{li2017holistic} proposed a framework with a human detector and a category-level segmentation module to segment the parts of objects at the instance level. These methods involve mutiple separate stages for instance localization, human parsing and result refinement. In comparison, the proposed NAN produces accurate multi-human parsing results through a single forward-pass in a time-efficient manner without tedious pre- or post-processing.

\begin{figure*}[t]
	\begin{center}
		\includegraphics[width=1\linewidth]{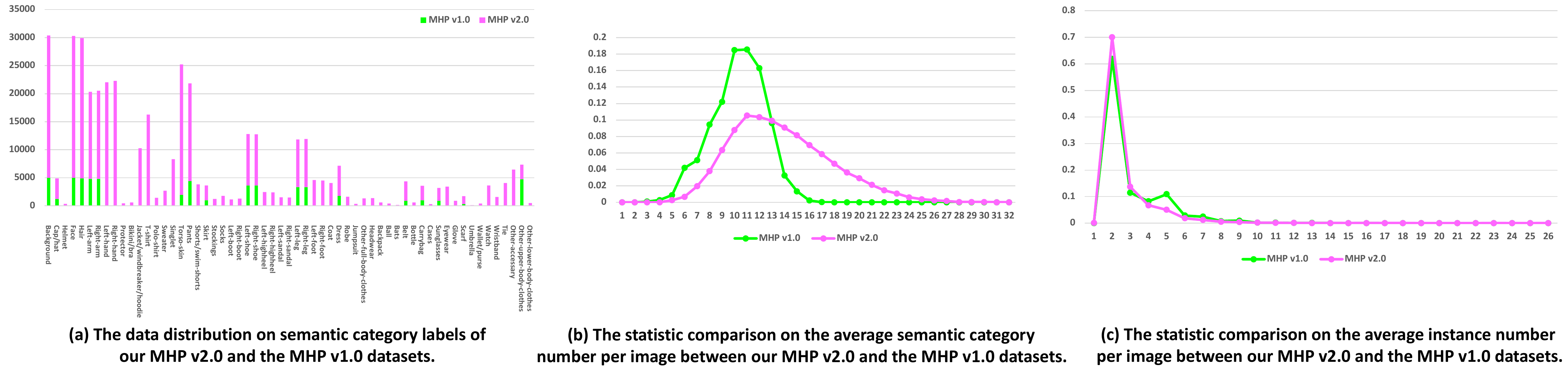}
	\end{center}
	\vspace{-4mm}
	\small
	\caption{\small Dataset statistics. Best viewed in color.}
	\label{fig: Fig3}
\end{figure*}

\section{Multi-Human Parsing Benchmark}
\label{Multi-Human Parsing Benchmark}

In this section, we introduce the ``\textbf{M}ulti-\textbf{H}uman \textbf{P}arsing (MHP v2.0)", a new large-scale dataset focusing on semantic understanding of humans in crowded scenes with several appealing properties.~1) It contains 25{,}403 elaborately annotated images with 58 fine-grained labels on body parts, fashion items and one background label, which is larger and more comprehensive than previous similar attempts~\cite{vineet2011human, li2017towards}.~2) The images within MHP v2.0 are collected from real-world scenarios, involving humans with various viewpoints, poses, occlusion,  interactions and resolution.~3) The background of images in MHP v2.0 is more complex and diverse than previous datasets. Some examples are showed in Fig.~\ref{fig: Fig2}. The MHP v2.0 dataset is expected to provide a new benchmark suitable for multi-human parsing together with a standard evaluation server where the test set will be kept secret to avoid overfitting.

\subsection{Image Collection and Annotation} 

\begin{figure}[t]
	\begin{center}
		\includegraphics[width=1\linewidth]{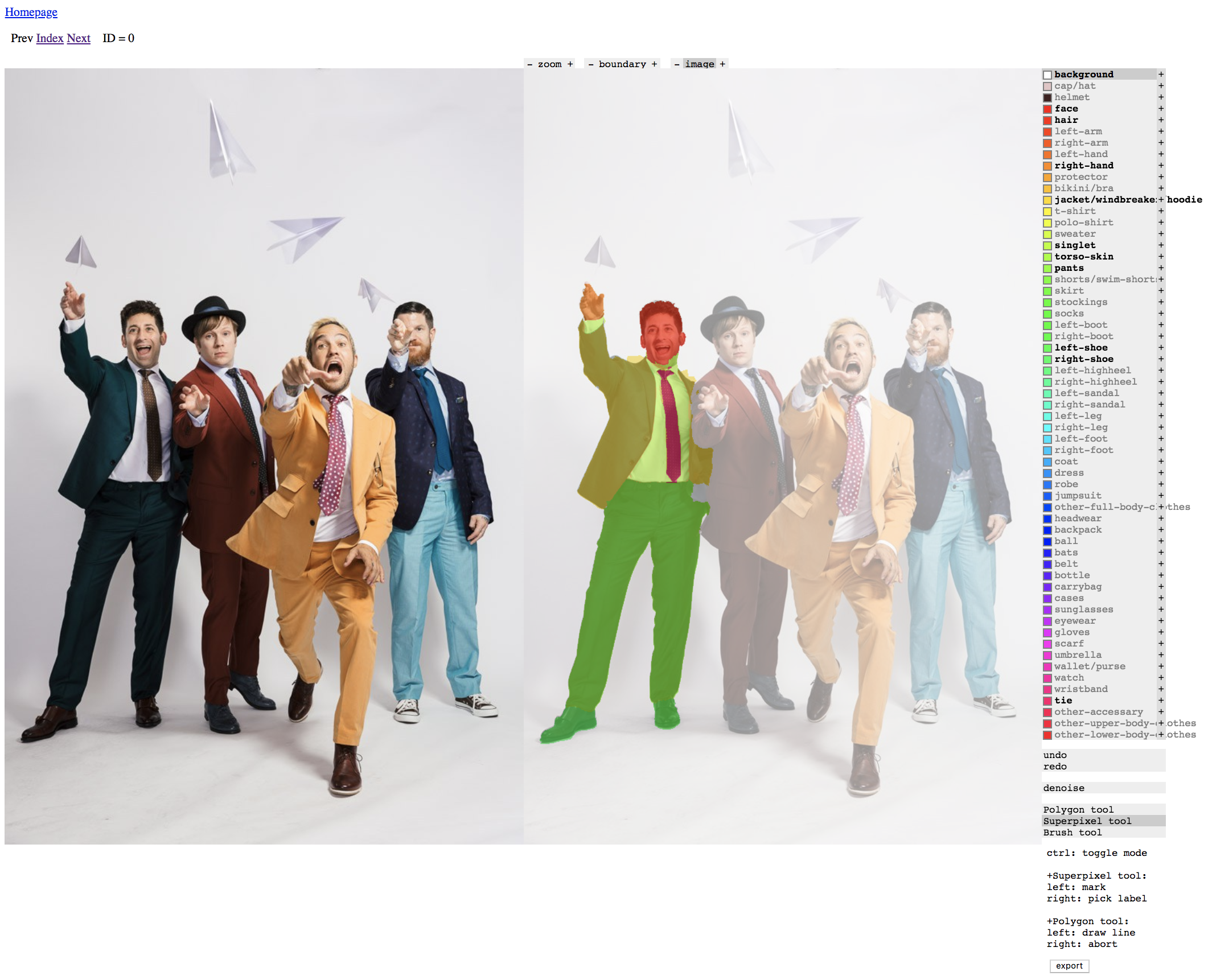}
	\end{center}
	\vspace{-4mm}
	\small
	\caption{\small Annotation tool for multi-human parsing. Best viewed in color.}
	\label{fig: Fig3_1}
	\vspace{-3mm}
\end{figure}

We manually specify some underlying relationships (such as family, couple, team, \emph{etc.}) and  possible scenes (such as sports, conferences, banquets, \emph{etc.}) to ensure the diversity of returned results. Based on any one of these specifications, corresponding multi-human images are located by performing Internet searches over Creative Commons licensed imagery. For each identified image, the contained human number and the corresponding URL are stored in a spreadsheet. Automated scrapping software is used to download the multi-human imagery and stores all relevant information in a relational database. Moreover, a pool of images containing clearly visible persons with interactions and rich fashion items is also constructed from the existing human-centric datasets~\cite{zhang2015beyond, chu2015multi, zhang2016facial, sapp2013modec, klare2015pushing}\footnote{PASCAL-VOC-2012~\cite{everingham2015pascal} and Microsoft COCO~\cite{lin2014microsoft} are not included due to limited percent of crowd-scene images with fine details of persons.} to augment and complement Internet scraping results.

After curating the imagery, manual annotation is conducted by professional data annotators, which includes two distinct tasks. The first task is manually counting the number of foreground persons and duplicating each image to several copies according to the count number. Each duplicated image is marked with the image ID, the contained person number and a self-index. The second is assigning the fine-grained pixel-wise label to each semantic category for each person instance. We implement an annotation tool and generate multi-scale superpixels of images based on~\cite{arbelaez2011contour} to speed up the annotation. See Fig.~\ref{fig: Fig3_1} for an example. Each multi-human image contains at least two instances. The annotation for each instance is done in a left-to-right order, corresponding to the duplicated image with the self-index from beginning to end. For each instance, 58 semantic categories are defined and annotated, including \emph{cap/hat}, \emph{helmet}, \emph{face}, \emph{hair}, \emph{left-arm}, \emph{right-arm}, \emph{left-hand}, \emph{right-hand}, \emph{protector}, \emph{bikini/bra}, \emph{jacket/windbreaker/hoodie}, \emph{t-shirt}, \emph{polo-shirt}, \emph{sweater}, \emph{singlet}, \emph{torso-skin}, \emph{pants}, \emph{shorts/swim-shorts}, \emph{skirt}, \emph{stockings}, \emph{socks}, \emph{left-boot}, \emph{right-boot}, \emph{left-shoe}, \emph{right-shoe}, \emph{left-highheel}, \emph{right-highheel}, \emph{left-sandal}, \emph{right-sandal}, \emph{left-leg}, \emph{right-leg}, \emph{left-foot}, \emph{right-foot}, \emph{coat}, \emph{dress}, \emph{robe}, \emph{jumpsuits}, \emph{other-full-body-clothes}, \emph{headwear}, \emph{backpack}, \emph{ball}, \emph{bats}, \emph{belt}, \emph{bottle}, \emph{carrybag}, \emph{cases}, \emph{sunglasses}, \emph{eyewear}, \emph{gloves}, \emph{scarf}, \emph{umbrella}, \emph{wallet/purse}, \emph{watch}, \emph{wristband}, \emph{tie}, \emph{other-accessaries}, \emph{other-upper-body-clothes} and \emph{other-lower-body-clothes}. Each instance has a complete set of annotations whenever the corresponding category appears in the current image. When annotating one instance, others are regarded as background. Thus, the resulting annotation set for each image consists of $N$ instance-level parsing masks, where $N$ is the number of persons in the image.

After annotation, manual inspection is performed on all images and corresponding annotations to verify the correctness. In cases where annotations are erroneous, the information is manually rectified by 5 well informed analysts. The whole work took around three months to accomplish by 25 professional data annotators.

\subsection{Dataset Splits and Statistics}

In total, there are 25{,}403 images in the MHP v2.0 dataset. Each image contains 2-26 person instances, with 3 on average. The resolution of the images ranges from 85$\times$100 to 4{,}511$\times$6{,}919, with 644$\times$718 on average. We spit the images into training, validation and testing sets. Following random selection, we arrive at a unique split consisting of 15{,}403 training and 5{,}000 validation images with publicly available annotations, as well as  5{,}000 testing images with annotations withheld for benchmarking purpose.  

The statistics \emph{w.r.t.} data distribution on 59 semantic categories, the average semantic category number per image and the average instance number per image in the MHP v2.0 dataset are illustrated in Fig.~\ref{fig: Fig3} (a), (b) and (c), respectively. In general, \emph{face}, arms and legs are the most remarkable parts of a human body. However, understanding humans in crowded scenes needs to analyze fine-grained details of each person of interest, including different body parts, clothes and accessaries. We therefore define 11 body parts, and 47 clothes and accessaries. Among these 11 body parts, we divide arms, hands, legs and feet into left and right side for more precise analysis, which also increases the difficulty of the task. We define \emph{hair}, \emph{face} and \emph{torso-skin} as the remaining three body parts, which can be used as auxiliary guidance for more comprehensive instance-level analysis. As for clothing categories, we have common clothes like \emph{coat}, \emph{jacket/windbreaker/hoodie}, \emph{sweater}, \emph{singlet}, \emph{pants}, \emph{shorts/swim-shorts} and shoes, confusing categories such as \emph{t-shirt} \emph{v.s.} \emph{polo-shirt}, \emph{stockings} \emph{v.s.} \emph{socks}, \emph{skirt} \emph{v.s.} \emph{dress} and \emph{robe}, and boots \emph{v.s.} sandals and highheels, and infrequent categories such as \emph{cap/hat}, \emph{helmet}, \emph{protector}, \emph{bikini/bra}, \emph{jumpsuits}, \emph{gloves} and \emph{scarf}. Furthermore, accessaries like \emph{sunglasses}, \emph{belt}, \emph{tie}, \emph{watch} and bags are also taken into account, which are common but hard to predict, especially for the small-scale ones. 

To summarize, the pre-defined semantic categories of MHP v2.0 involve most body parts, clothes and accessaries of different styles for men, women and children in all seasons. The images in the MHP v2.0 dataset contain diverse instance numbers, viewpoints, poses, occlusion, interactions and background complexities. MHP v2.0 aligns better with real-world scenarios and serves as a more realistic benchmark for human-centric analysis, which pushes the frontiers of fine-grained multi-human parsing research.

\section{Deep Nested Adversarial Networks}

\begin{figure*}[t]
	\begin{center}
		\includegraphics[width=1\linewidth]{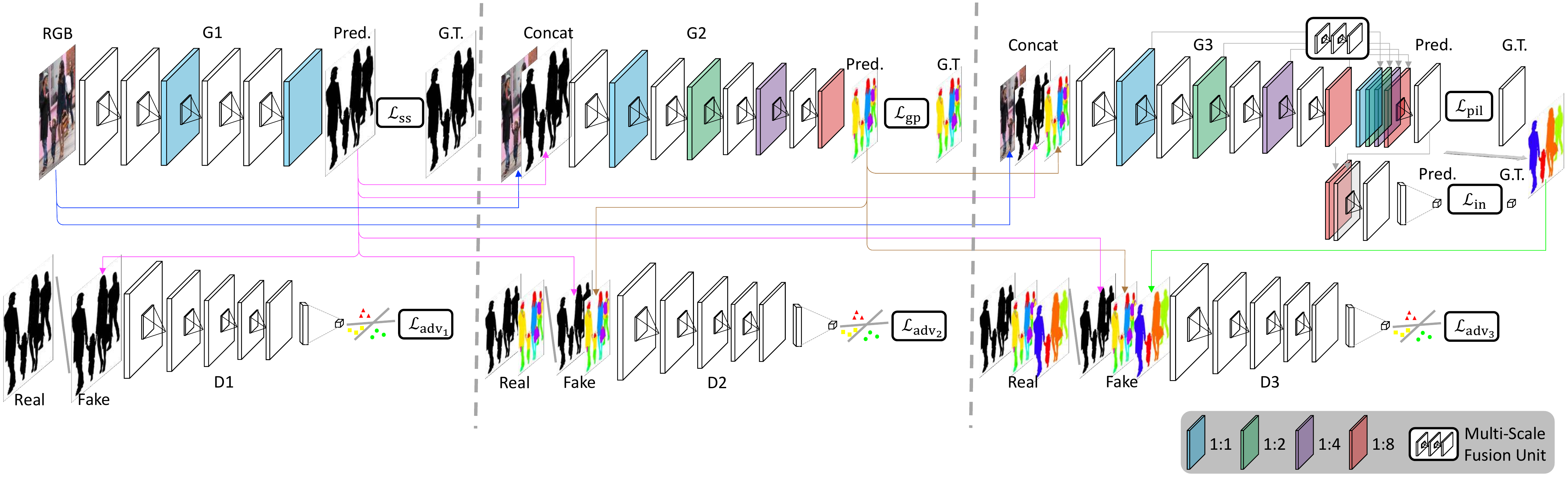}
	\end{center}
	\vspace{-4mm}
	\small
	\caption{\small Deep Nested Adversarial Networks (NAN) for multi-human parsing. NAN consists of three GAN-like sub-nets, respectively performing semantic saliency prediction, instance-agnostic parsing and instance-aware clustering. Each sub-task is simpler than the original multi-human parsing task, and is more easily addressed by the corresponding sub-net. The sub-nets depend on each other, forming a causal nest by dynamically boosting each other via an adversarial strategy. Such a structure enables effortless gradient \textbf{}Back\textbf{P}ropagation (BP) of NAN such that it can be trained in a holistic, end-to-end way. NAN produces accurate multi-human parsing results through a single forward-pass in a time-efficient manner without tedious pre- or post-processing. Best viewed in color.}
	\label{fig: Fig4}
\end{figure*}

As shown in Fig.~\ref{fig: Fig4}, the proposed deep \textbf{N}ested \textbf{A}dversarial \textbf{N}etwork (NAN) model consists of three GAN-like sub-nets that  jointly perform semantic saliency prediction, instance-agnostic parsing and instance-aware clustering end-to-end. NAN produces accurate multi-human parsing results through a single forward-pass in a time-efficient manner without tedious pre- or post-processing. We now present each component in details.

\subsection{Semantic Saliency Prediction}

Large modality and interaction variations are the main challenge to multi-human parsing and also the key obstacle to learning a well-performing human-centric analysis model. To address this problem, we propose to decompose the original task into three granularities and adaptively impose a prior on the specific process, each with the aid of a GAN-based sub-net. This reduces the training complexity and leads to better empirical performance with limited data. 

The first sub-net estimates semantic saliency maps to locate the most noticeable and eye-attracting human regions in images, which serves as a basic prior to facilitate further processing on humans, as illustrated in Fig.~\ref{fig: Fig4} left. We formulate semantic saliency prediction as a binary pixel-wise labelling problem to segment out foreground \emph{v.s.} background. Inspired by the recent success of \textbf{F}ully \textbf{C}onvolutional \textbf{N}etwork\textbf{s} (FCNs)~\cite{long2015fully} based methods on image-to-image applications \cite{li2017instance, he2017mask}, we leverage an FCN backbone (FCN-8s~\cite{long2015fully}) as the generator $\mathrm{G1}_{\theta_1} : \mathbb{R}^{H \times W \times C} \mapsto \mathbb{R}^{H \times W \times C{'}}$ of NAN for semantic saliency prediction, where $\theta_1$ denotes the network parameters, and $H$, $W$, $C$ and $C{'}$ denote the image height, width, channel number and semantic category (\emph{i.e.}, foreground plus background) number, repectively.

Formally, let the input RGB image be denoted by $x$ and the semantic saliency map be denoted by $x{'}$, then
\begin{equation}
\small
x{'} := \mathrm{G1}_{\theta_1}(x).
\end{equation}

The key requirements for $\mathrm{G1}$ are that the semantic saliency map $x{'}$ should present indistinguishable properities compared with a real one (\emph{i.e.}, ground truth) in appearance while preserving the intrinsic contextually remarkable information.

To this end, we propose to learn $\theta_1$ by minimizing a combination of two losses:
\begin{equation}
\small
\mathcal{L}_{\mathrm{G1_{\theta_1}}}=-\lambda_1 \mathcal{L}_{\mathrm{adv}_1}+\lambda_2\mathcal{L}_{\mathrm{ss}},
\end{equation}
where $\mathcal{L}_{\mathrm{adv}_1}$ is the \textbf{adv}ersarial loss for refining realism and alleviating artifacts, $\mathcal{L}_{\mathrm{ss}}$ is the \textbf{s}emantic \textbf{s}aliency loss for pixel-wise image labelling, $\lambda$ are weighting parameters among different losses.

$\mathcal{L}_{\mathrm{ss}}$ is a pixel-wise cross-entropy loss calculated based on the binary pixel-wise annotations to learn $\theta_1$:
\begin{equation}
\small
\mathcal{L}_{\mathrm{ss}} = \mathcal{L}_{\mathrm{ss}}(X{'}(\theta_1)|X).
\end{equation}

$\mathcal{L}_{\mathrm{adv}_1}$ is proposed to narrow the gap between the distributions of generated and real results. To facilitate this process, we leverage a \textbf{C}onvolutional \textbf{N}eural \textbf{N}etwork (CNN) backbone as the discriminator $\mathrm{D1}_{\phi_1}: \mathbb{R}^{H \times W \times C{'}} \mapsto \mathbb{R}^{1}$ to be as simple as possible to avoid typical GAN tricks. We alternatively optimize $\mathrm{G1_{\theta_1}}$ and $\mathrm{D1}_{\phi_1}$ to learn $\theta_1$ and $\phi_1$:
\begin{equation}
\small
\begin{cases}
\mathcal{L}_{\mathrm{adv}_1}^{\mathrm{G1}} = \mathcal{L}_{\mathrm{adv}_1}(K(\theta_1)|X{'}(\theta_1), X{'}_{\mathrm{GT}}), \\
\mathcal{L}_{\mathrm{adv}_1}^{\mathrm{D1}} = \mathcal{L}_{\mathrm{adv}_1}(K(\phi_1)|X{'}(\theta_1), X{'}_{\mathrm{GT}}),
\end{cases}
\end{equation}
where $K$ denotes the binary real \emph{v.s.} fake indicator.

\subsection{Instance-Agnostic Parsing}

The second sub-net concatenates the information from the original RGB image with semantic saliency prior as input and estimates a fine-grained instance-agnostic parsing map, which further serves as stronger semantic guidance from the global perspective to facilitate instance-aware clustering, as illustrated in Fig.~\ref{fig: Fig4} middle. We formulate instance-agnostic parsing as a multi-class dense classification problem to mask semantically consistent regions of body parts and fashion items. Inspired by the leading performance of the skip-net on recognition tasks~\cite{wu2016wider, he2016deep}, we modify a skip-net (WS-ResNet~\cite{wu2016wider}) into an FCN-based architecture as the generator $\mathrm{G_2}_{\theta_2} : \mathbb{R}^{H \times W \times (C+C{'})} \mapsto \mathbb{R}^{H/8 \times W/8 \times C{''}}$ of NAN to learn a highly non-linear transformation for instance-agnostic parsing, where $\theta_2$ denotes the network parameters for the generator and $C{''}$ denotes the semantic category number. The prediction is downsampled by $8$ for accuracy \emph{v.s.} speed trade-off. Contextual information from global and local regions compensates each other and naturally benefits human parsing. The hierarchical features within a skip-net are multi-scale in nature due to the increasing receptive field sizes, which are combined together via skip connections. Such a combined representation comprehensively maintains the contextual information, which is crucial for generating smooth and accurate parsing results.

Formally, let the instance-agnostic parsing map be denoted by $x{''}$, then
\begin{equation}
\small
x{''} := \mathrm{G2}_{\theta_2}(x \cup x{'}).
\end{equation}

Similar to the first sub-net, we propose to learn $\theta_2$ by minimizing:
\begin{equation}
\small
\mathcal{L}_{\mathrm{G2_{\theta_2}}}=-\lambda_3 \mathcal{L}_{\mathrm{adv}_2}+\lambda_4\mathcal{L}_{\mathrm{gp}},
\end{equation}
where $\mathcal{L}_{\mathrm{gp}}$ is the \textbf{g}lobal \textbf{p}arsing loss for semantic part labelling.

$\mathcal{L}_{\mathrm{gp}}$ is a standard pixel-wise cross-entropy loss calculated based on the multi-class pixel-wise annotations to learn $\theta_2$. $\theta_1$ is also slightly finetuned due to the hinged gradient backpropagation route within the nested structure:
\begin{equation}
\small
\mathcal{L}_{\mathrm{gp}} = \mathcal{L}_{\mathrm{gp}}(X{''}(\theta_2, \theta_1)|X \cup X{'}(\theta_1)).
\end{equation}

$\mathcal{L}_{\mathrm{adv}_2}$ is proposed to ensure the correctness and realism of the current phase and also the previous one for information flow consistency. To facilitate this process, we leverage a same CNN backbone with $\mathrm{D1}_{\phi_1}$ as the discriminator $\mathrm{D2}_{\phi_2}: \mathbb{R}^{H \times W \times (C{'}+C{''})} \mapsto \mathbb{R}^{1}$, which are learned separately. We alternatively optimize $\mathrm{G2_{\theta_2}}$ and $\mathrm{D2}_{\phi_2}$ to learn $\theta_2$, $\phi_2$ and slightly finetune $\theta_1$:
\begin{equation}
\small
\begin{cases}
\mathcal{L}_{\mathrm{adv}_2}^{\mathrm{G2}} = \mathcal{L}_{\mathrm{adv}_2}(K(\theta_2)|X{'}(\theta_1) \cup X{''}(\theta_2, \theta_1), X{'}_{\mathrm{GT}} \cup X{''}_{\mathrm{GT}}), \\
\mathcal{L}_{\mathrm{adv}_2}^{\mathrm{D2}} = \mathcal{L}_{\mathrm{adv}_2}(K(\phi_2)|X{'}(\theta_1) \cup X{''}(\theta_2, \theta_1), X{'}_{\mathrm{GT}} \cup X{''}_{\mathrm{GT}}).
\end{cases}
\end{equation}

\subsection{Instance-Aware Clustering}

The third sub-net concatenates the information from the original RGB image with semantic saliency and instance-agnostic parsing priors as input and estimates an instance-aware clustering map by associating each semantic parsing mask to one of the person instances in the scene, as illustrated in Fig.~\ref{fig: Fig4} right. Inspired by the observation that a human glances at an image and instantly knows how many and where the objects are in the image, we formulate instance-aware clustering by parallelly inferring the instance number and pixel-wise instance location, discarding the requirement of time-consuming region proposal generation. We modify a same backbone architecture $\mathrm{G2}_{\theta_2}$ to incorporate two sibling branches as the generator $\mathrm{G_3}_{\theta_3} : \mathbb{R}^{H/8 \times W/8 \times (C+C{'}+C{''})} \mapsto \mathbb{R}^{H/8 \times W/8 \times M} \cup \mathbb{R}^1$ of NAN for location-sensitive learning, where $\theta_3$ denotes the network parameters for the generator and $M$  denotes the pre-defined instance location coordinate number. As multi-scale features integrating both global and local contextual information are crucial for increasing location prediction accuracy, we further augment the pixel-wise instance location prediction branch with a \textbf{M}ulti-\textbf{S}cale \textbf{F}usion \textbf{U}nit (MSFU) to fuse shallow-, middle- and deep-level features, while using the feature maps downsampled by $8$ concatenated with feature maps from the first branch for instance number regression.

Formally, let the pixel-wise instance location map be denoted by $\tilde{x}$ and the instance number be denoted by $n$, then
\begin{equation}
\small
\tilde{x} \cup n := \mathrm{G3}_{\theta_3}(x \cup x{'} \cup x{''}).
\end{equation}

We propose to learn $\theta_3$ by minimizing:
\begin{equation}
\small
\mathcal{L}_{\mathrm{G3_{\theta_3}}}=-\lambda_5 \mathcal{L}_{\mathrm{adv}_3}+\lambda_6\mathcal{L}_{\mathrm{pil}}+\lambda_7\mathcal{L}_{\mathrm{in}},
\end{equation}
where $\mathcal{L}_{\mathrm{pil}}$ is the \textbf{p}ixel-wise \textbf{i}nstance \textbf{l}ocation loss for pixel-wise instance location regression and $\mathcal{L}_{\mathrm{in}}$ is the \textbf{i}nstance \textbf{n}umber loss for instance number regression.

$\mathcal{L}_{\mathrm{pil}}$ is a standard smooth-$\ell_1$ loss~\cite{girshick2015fast} calculated based on the foreground pixel-wise instance location annotations to learn $\theta_3$. Since a person instance can be identified by its top-left corner $(x^l, y^l)$ and bottom-right corner $(x^r, y^r)$ of the surrounding bounding box, for each pixel belonging to the person instance, the pixel-wise instance location vector is defined as $[x^l/w, y^l/h, x^r/w, y^r/h]$, where $w$ and $h$ are the width and height of the person instance for normalization, respectively. $\mathcal{L}_{\mathrm{in}}$ is a standard $\ell_2$ loss calculated based on the instance number annotations to learn $\theta_3$. $\theta_2$ and $\theta_1$ are also slightly finetuned due to the chained schema within the nest:
\begin{equation}
\small
\begin{cases}
\mathcal{L}_{\mathrm{pil}} = \mathcal{L}_{\mathrm{pil}}(\tilde{X}(\theta_3, \theta_2, \theta_1)|X \cup X{'}(\theta_1) \cup X{''}(\theta_2, \theta_1)), \\
\mathcal{L}_{\mathrm{in}} = \mathcal{L}_{\mathrm{in}}(N(\theta_3, \theta_2, \theta_1)|X \cup X{'}(\theta_1) \cup X{''}(\theta_2, \theta_1)).
\end{cases}
\end{equation}

Given these information, instance-aware clustering maps can be effortlessly generated with little computational overhead, which are denoted by $\hat{X}\in \mathbb{R}^{M{'}}$. Similar to $\mathcal{L}_{\mathrm{adv}_2}$, $\mathcal{L}_{\mathrm{adv}_3}$ is proposed to ensure the correctness and realism of all phases for the information flow consistency. To facilitate this process, we leverage a same CNN backbone with $\mathrm{D2}_{\phi_2}$ as the discriminator $\mathrm{D3}_{\phi_3}: \mathbb{R}^{H \times W \times (C{'}+C{''}+M{'})} \mapsto \mathbb{R}^{1}$, which are learned separately. We alternatively optimize $\mathrm{G3_{\theta_3}}$ and $\mathrm{D3}_{\phi_3}$ to learn $\theta_3$, $\phi_3$ and slightly finetune $\theta_2$ and $\theta_1$:
\begin{equation}
\tiny
\begin{cases}
\mathcal{L}_{\mathrm{adv}_3}^{\mathrm{G3}} = \mathcal{L}_{\mathrm{adv}_3}(K(\theta_3)|X{'}(\theta_1) \cup X{''}(\theta_2, \theta_1) \cup \hat{X}(\theta_3, \theta_2, \theta_1), X{'}_{\mathrm{GT}} \cup X{''}_{\mathrm{GT}} \cup \hat{X}_{\mathrm{GT}}), \\
\mathcal{L}_{\mathrm{adv}_3}^{\mathrm{D3}} = \mathcal{L}_{\mathrm{adv}_3}(K(\phi_3)|X{'}(\theta_1) \cup X{''}(\theta_2, \theta_1) \cup \hat{X}(\theta_3, \theta_2, \theta_1), X{'}_{\mathrm{GT}} \cup X{''}_{\mathrm{GT}} \cup \hat{X}_{\mathrm{GT}}).
\end{cases}
\end{equation}

\subsection{Training and Inference}

The goal of NAN is to use sets of real targets to learn three GAN-like sub-nets that mutually boost and jointly accomplish multi-human parsing. Each separate loss serves as a deep supervision within the nested structure benefitting network convergence. The overall objective function for NAN is
\begin{equation}
\small
\mathcal{L}_{\mathrm{NAN}}=-\sum\limits_{i=1}^{3}\lambda_i \mathcal{L}_{\mathrm{adv}_i}+\lambda_4\mathcal{L}_{\mathrm{ss}}+\lambda_5\mathcal{L}_{\mathrm{gp}}+\lambda_6\mathcal{L}_{\mathrm{pil}}+\lambda_7\mathcal{L}_{\mathrm{in}}.
\end{equation}

Clearly, the NAN is end-to-end trainable and can be optimized with the proposed nested adversarial learning strategy and BP algorithm. 

During testing, we simply feed the input image $X$ into NAN to get the instance-agnostic parsing map $X{''}$ from  $\mathrm{G2}_{\theta_2}$, pixel-wise instance location map $\tilde{X}$ and instance number $N$ from $\mathrm{G3}_{\theta_3}$. Then we employ an off-the-shelf clustering method~\cite{ng2002spectral} to obtain the instance-aware clustering map $\hat{X}$. Example results are visualized in Fig.~\ref{fig: Fig5_1}.

\section{Experiments}

We evaluate NAN qualitatively and quantitatively under various settings and granularities for understanding humans in crowded scenes. In particular,
we evaluate multi-human parsing performance on the MHP v2.0 dataset proposed in this work, as well as the MHP v1.0~\cite{li2017towards} and PASCAL-Person-Part~\cite{chen2014detect} benchmark datasets. We also evaluate instance-agnostic parsing and instance-aware clustering results on the Buffy~\cite{vineet2011human} benchmark dataset, which are byproducts of NAN.

\subsection{Experimental Settings}
\label{sec5.1}

\subsubsection{Implementation Details}

Throughout the experiments, the sizes of the RGB image $X$, the semantic saliency prediction $X{'}$, inputs to the discriminator $\mathrm{D1}_{\phi_1}$ and inputs to the generator $\mathrm{G2}_{\theta_2}$ are fixed as $512 {\times} 512$; the sizes of the instance-agnostic parsing prediction $X{''}$, instance-aware clustering prediction $\hat{X}$, inputs to the discriminator $\mathrm{D2}_{\phi_2}$, inputs to the generator $\mathrm{G3}_{\theta_3}$, inputs to the discriminator $\mathrm{D3}_{\phi_3}$ and instance location map $\tilde{X}$ are fixed as $64 {\times} 64$; the channel number of the pixel-wise instance location map is fixed as $4$, incorporating two corner points of the associated bounding box; the constraint factors $\lambda_i, i\in \{1,2,3,4,5,6,7\}$ are empirically fixed as $0.01, 0.01, 0.01, 1.00, 1.00, 10.00$ and $1.00$, respectively; the generator $\mathrm{G1}_{\theta_1}$ is initialized with FCN-8s~\cite{long2015fully} by replacing the last layer with a new convolutional layer with kernel size $1\times 1\times 2$, pretrained on PASCAL-VOC-2011~\cite{pascal-voc-2011} and finetuned on the target dataset; the generator $\mathrm{G2}_{\theta_2}$ is initialized with WS-ResNet~\cite{wu2016wider} by eliminating the spatial pooling layers, increasing the strides of the first convolutional layers up to 2 in B$_i, i\in \{2,3,4\}$, eliminating the top-most global pooling layer and the linear classifier, and adding two new convolutional layers with kernel sizes $3\times 3\times 512$ and $1\times 1\times C{''}$, pretrained on ImageNet~\cite{russakovsky2015imagenet} and PASCAL-VOC-2012~\cite{everingham2015pascal}, and finetuned on the target dataset; the generator $\mathrm{G3}_{\theta_3}$ is initialized with the same backbone architecture and pre-trained weights with $\mathrm{G2}_{\theta_2}$ (which are learned separately), by further augmenting it with two sibling branches for pixel-wise instance location map prediction and instance number prediction, where the first branch utilizes a MSFU (three convolutional layers with kernal sizes $3\times 3\times i, i\in \{128,128,4\}$ for specific scale adaption) ended with a convolutional layer with kernel size $1\times 1\times 4$ for multi-scale feature aggregation and a final convolutional layer with kernel size $1\times 1\times 4$ for location regression and the second branch utilizes the feature maps downsampled by 8 concatenated with the feature maps from the first branch ended with a global pooling layer, a hidden 512-way fully-connected layer and a final 1-way fully-connected layer for instance number regression; the three discriminators $\mathrm{Di}_{\phi_i}, i\in \{1,2,3\}$ (which are learned separately) are all initialized with a VGG-16~\cite{simonyan2014very} by adding a new convolutional layer at the very begining with kernel size $1\times 1\times 3$ for input adaption, and replacing the last layer with a new 1-way fully-connected layer activated by sigmoid, pre-trained on ImageNet~\cite{russakovsky2015imagenet} and finetuned on the target dataset; the newly added layers are randomly initialized by drawing weights from a zero-mean Gaussian distribution with standard deviation $0.01$; we employ an off-the-shelf clustering method~\cite{ng2002spectral} to obtain the instance-aware clustering map $\hat{X}$; the dropout ratio is empirically fixed as $0.7$; the weight decay and batch size are fixed as $5{\times} 10^{-3}$ and $4$, respectively;  We use an initial learning rate of $1{\times} 10^{-6}$ for pre-trained layers, and $1{\times} 10^{-4}$ for newly added layers in all our experiments; we decrease the learning rate to $1/10$ of the previous one after 20 epochs and train the network for roughly 60 epochs one after the other; the proposed network is implemented based on the publicly available TensorFlow~\cite{abadi2016tensorflow} platform, which is trained using Adam ($\beta_1{=}0.5$) on four NVIDIA GeForce GTX TITAN X GPUs with 12G memory; the same training setting is utilized for all our compared network variants; we evaluate the testing time by averaging the running time for images on the target set on NVIDIA GeForce GTX TITAN X GPU and Intel Core i7-4930K CPU@3.40GHZ; our NAN can rapidly process one $512\times 512$ image in about 1 second, which compares much favorably to other state-of-the-art approaches, as the current state-of-the-art methods~\cite{li2017towards, jiang2016detangling, li2017holistic} rely on region proposal preprocessing and complex processing steps.

\vspace{-2mm}
\subsubsection{Evaluation Metrics}

Following~\cite{li2017towards}, we use the \textbf{A}verage \textbf{P}recision based on \textbf{p}art ($\mathrm{AP}^{p}$) and \textbf{P}ercentage of \textbf{C}orrectly parsed semantic \textbf{P}arts (PCP) metrics for multi-human parsing evaluation. Different from the \textbf{A}verage \textbf{P}recision based on \textbf{r}egion ($\mathrm{AP}^{r}$) used in instance segmentation~\cite{liang2015proposal, hariharan2014simultaneous}, $\mathrm{AP}^{p}$ uses part-level pixel \textbf{I}ntersection \textbf{o}ver \textbf{U}nion (IoU) of different semantic part categories within a person instance to determine if one instance is a true positive. We prefer $\mathrm{AP}^{p}$ over $\mathrm{AP}^{r}$ as we focus on human-centric analysis and we aim to investigate to how well a person instance as a whole is parsed. Additionally, we also report the $\mathrm{AP}^{p}_{vol}$, which is the mean of the $\mathrm{AP}^{p}$ at IoU thresholds ranging from $0.1$ to $0.9$, in increments of 0.1. As $\mathrm{AP}^{p}$ averages the IoU of each semantic part category, it fails to reflect how many semantic parts are correctly parsed. We further incorporate the PCP, originally used in human pose estimation~\cite{ferrari2008progressive, chen2014detect}, to evaluate the parsing quality within person instances. For each true-positive person instance, we find all the semantic categories (excluding background) with pixel IoU larger than a threshold, which are regarded as correctly parsed. The PCP of one person instance is the ratio between the correctly parsed semantic category number and the total semantic category number of that person. Missed person instances are assigned with $0$ PCP. The overall PCP is the average PCP for all person instances. Note that PCP is also a human-centric evaluation metric.

\subsection{Evaluations on the MHP v2.0 Benchmark}

The MHP v2.0 dataset proposed in this paper is the largest and most comprehensive multi-human parsing benchmark to date, which extends MHP v1.0~\cite{li2017towards} to push the frontiers of understanding humans in crowded scenes by containing 25{,}403 elaborately annotated images with 58 fine-grained semantic category labels. Annotation examples are visualized in Fig.~\ref{fig: Fig2} (c). The data are randomly organized into 3 splits, consisting of 15{,}403 training and 5{,}000 validation images with publicly available annotations, as well as 5{,}000 testing images with annotations withheld for benchmarking purpose. Evaluation systems report the $\mathrm{AP}^{p}$ and PCP over the validation and testing sets.

\vspace{-2mm}
\subsubsection{Component Analysis}

We first investigate different architectures and loss function combinations of NAN to see their respective roles in multi-human parsing. We compare 16 variants from four aspects, \emph{i.e.}, different baselines (Mask R-CNN~\cite{he2017mask}\footnote{As existing instance segmentation methods only offer silhouettes of different person instances, for comparison, we combine them with our instance-agnostic parsing prediction to generate the final multi-human parsing results.} and MH-Parser~\cite{li2017towards}), different network structures (w/o $\mathrm{G1}$, $\mathrm{G2}$ w/o concatenated input (RGB only), $\mathrm{G3}$ w/o concatenated input (RGB only), w/o $\mathrm{D1}$, w/o $\mathrm{D2}$, $\mathrm{D2}$ w/o concatenated input, w/o $\mathrm{D3}$, $\mathrm{D3}$ w/o concatenated input, w/o MSFU), our proposed NAN, and upperbounds ($X{'}_\mathrm{GT}$: use the ground truth semantic saliency maps instead of $\mathrm{G1}$ prediction while keeping other settings the same; $X{''}_\mathrm{GT}$: use the ground truth instance-agnostic parsing maps instead of $\mathrm{G2}$ prediction while keeping other settings the same; $N_\mathrm{GT}$: use the ground truth instance number instead of $\mathrm{G3}$ prediction while keeping other settings the same; $\tilde{X}_\mathrm{GT}$: use the ground truth pixel-wise instance location maps instead of $\mathrm{G3}$ prediction while keeping other settings the same).

\begin{table}[t]
	\newcommand{\tabincell}[2]{\begin{tabular}{@{}#1@{}}#2\end{tabular}}
	\tiny
	\caption{\small Component analysis on the MHP v2.0 validation set.}
	\vspace{-5mm}
	\begin{center}
		\begin{tabular}{ccccc}
			\hline
			\tabincell{c}{Setting} & \tabincell{c}{Method} & \tabincell{c}{$\mathrm{AP}^{p}_{0.5} (\%)$} & \tabincell{c}{$\mathrm{AP}^{p}_{vol} (\%)$} & \tabincell{c}{PCP$_{0.5} (\%)$} \\
			\hline
			\multirow{3}{*}{Baseline} & Mask R-CNN~\cite{he2017mask} & 14.50 & 33.51 & 25.12 \\
			& MH-Parser~\cite{li2017towards} & 18.05 & 35.87 & 26.91 \\
			\hline
			\multirow{9}{*}{Network Structure} & w/o $\mathrm{G1}$ & 22.67 & 38.11 & 31.95 \\
			& $\mathrm{G2}$ w/o concatenated input & 21.88 & 36.79 & 29.02 \\
			& $\mathrm{G3}$ w/o concatenated input & 22.36 & 35.92 & 25.48 \\
			& w/o $\mathrm{D1}$ & 23.81 & 33.95 & 27.59 \\
			& w/o $\mathrm{D2}$ & 19.02 & 29.66 & 22.89 \\
			& $\mathrm{D2}$ w/o concatenated input & 21.55 & 31.94 & 24.90 \\
			& w/o $\mathrm{D3}$ & 20.62 & 32.83 & 26.22 \\
			& $\mathrm{D3}$ w/o concatenated input & 21.80 & 34.54 & 27.30 \\
			& w/o MSFU & 18.76 & 26.62 & 24.94 \\
			\hline
			Ours & NAN & 24.83 & 42.77 & 34.37 \\
			\hline
			\multirow{4}{*}{Upperbound} & $X{'}_\mathrm{GT}$ & 26.17 & 43.59 & 38.11 \\
			& $X{''}_\mathrm{GT}$ & 28.98 & 48.55 & 38.03 \\
			& $N_\mathrm{GT}$ & 28.39 & 47.76 & 39.25 \\
			& $\tilde{X}_\mathrm{GT}$ & 30.18 & 51.44 & 41.18 \\
			\hline
		\end{tabular}
	\end{center}
	\label{tab: Tab2}
	\vspace{-6mm}
\end{table}

The performance comparison in terms of $\mathrm{AP}^{p}$@IoU=0.5, $\mathrm{AP}^{p}_{vol}$ and PCP@IoU=0.5 on the MHP v2.0 validation set is reported in Tab.~\ref{tab: Tab2}. By comaring the results from the $1^{st}$ \emph{v.s.} $3^{rd}$ panels, we observe that our proposed NAN consistently outperforms the baselines Mask R-CNN~\cite{he2017mask} and MH-Parser~\cite{li2017towards} by a large margin, \emph{i.e.}, $10.33\%$ and $6.78\%$ in terms of $\mathrm{AP}^{p}$, $9.26\%$ and $6.90\%$ in terms of $\mathrm{AP}^{p}_{vol}$, and $9.25\%$ and $7.46\%$ in terms of PCP. Mask R-CNN~\cite{he2017mask} suffers difficulties to differentiate entangled humans. MH-Parser~\cite{li2017towards} involves multiple stages for instance localization, human parsing and result refinement with high complexity, yielding sub-optimal results, whereas NAN parses semantic categories, differentiates different person instances and refines results simultaneously through deep nested adversarial learning in an effective yet time-efficient manner. By comaring the results from the $2^{nd}$ \emph{v.s.} $3^{rd}$ panels, we observe that NAN consistently outperforms the 9 variants in terms of network structure. In particular, w/o $\mathrm{G1}$ refers to truncating the semantic saliency prediction sub-net from NAN, leading to $2.16\%$, $4.66\%$ and $2.42\%$ performance drop in terms of all metrics. This verifies the necessity of semantic saliency prediction that locates the most noticeable human regions in images to serve as a basic prior to facilitate further human-centic processing. The superiority of incorporating adaptive prior information to specific process can be verified by comparing $\mathrm{Gi}, i\in \{2,3\}$ w/o concatenated input with NAN, \emph{i.e.}, $2.95\%$, $5.98\%$ and $5.35\%$; $2.47\%$, $6.85\%$ and $8.89\%$ differences in terms of all metrics. The superiority of incorporating adversarial learning to specific process can be verified by comparing w/o $\mathrm{Di}, i\in \{1,2,3\}$ with NAN, \emph{i.e.}, $1.02\%$, $8.82\%$ and $6.78\%$; $5.81\%$, $13.11\%$ and $11.48\%$; $4.21\%$, $9.94\%$ and $8.15\%$ decrease in terms of all metrics. Nested adversarial learning strategy ensures the correctness and realism of all phases for information flow consistency, the superiority of which is verified by comparing $\mathrm{Di}, i\in \{2,3\}$ w/o concatenated input with NAN, \emph{i.e.}, $3.28\%$, $10.83\%$ and $9.47\%$; $3.03\%$, $8.23\%$ and $7.07\%$ decline in terms of all metrics. MSFU dynamically fuses multi-scale features for enhancing instance-aware clustering accuracy, the superiority of which is verified by comparing w/o MSFU with NAN, \emph{i.e.}, $6.07\%$, $16.15\%$ and $9.43\%$ drop in terms of all metrics. Finally, we also evaluate the limitations of our current algorithm. By comparing $X{'}_\mathrm{GT}$ with NAN, only $1.34\%$, $0.82\%$ and $3.74\%$ improvement in term of all metrics are obtained, which shows that the errors from semantic saliency prediction are already small and have only little effect on the final results. A large gap between $28.98\%$, $48.55\%$ and $38.03\%$ of $X{''}_\mathrm{GT}$ and $24.83\%$, $42.77\%$ and $34.37\%$ of NAN shows that a better instance-agnostic parsing network architecture can definitely help improve the performance of multi-human parsing under our NAN framework. By comparing $N_\mathrm{GT}$ and $\tilde{X}_\mathrm{GT}$ with NAN, $3.56\%$, $4.99\%$ and $4.88\%$; $5.35\%$, $8.67\%$ and $6.81\%$ improvement in term of all metrics are obtained, which shows that accurate instance-aware clustering results are critical for superior multi-human parsing. 

\subsubsection{Quantitative Comparison}

\begin{table*}[t]\setlength{\tabcolsep}{4pt}
	\scriptsize
	\caption{\small Multi-human parsing quantitative comparison on the MHP v2.0 testing set.}
	\vspace{-5mm}
	\begin{center}
		\begin{tabular}{ccccccccccc}
			\hline
			\multirow{2}{*}{Method} &\multicolumn{3}{c}{All} & \multicolumn{3}{c}{Inter$_{20\%}$} & \multicolumn{3}{c}{Inter$_{10\%}$} & \multirow{2}{*}{Speed (img/s)} \\
			\cline{2-10}
			& {$\mathrm{AP}^{p}_{0.5} (\%)$} & {$\mathrm{AP}^{p}_{vol} (\%)$} & {PCP$_{0.5} (\%)$} & {$\mathrm{AP}^{p}_{0.5} (\%)$} & {$\mathrm{AP}^{p}_{vol} (\%)$} & {PCP$_{0.5} (\%)$} & {$\mathrm{AP}^{p}_{0.5} (\%)$} & {$\mathrm{AP}^{p}_{vol} (\%)$} & {PCP$_{0.5} (\%)$} \\
			\hline
			Mask R-CNN~\cite{he2017mask} & 14.90 & 33.88 & 25.11 & 4.77 & 24.28 & 12.75 & 2.23 & 20.73 & 8.38 & - \\
			MH-Parser~\cite{li2017towards} & 17.99 & 36.08 & 26.98 & 13.38 & 34.25 & 22.31 & 13.25 & 34.29 & 21.28 & 14.94 \\
			\hline
			NAN & \textbf{25.14} & \textbf{41.78} & \textbf{32.25} & \textbf{18.61} & \textbf{38.90} & \textbf{27.93} & \textbf{14.88} & \textbf{37.49} & \textbf{24.61} & \textbf{1.08} \\
			\hline
		\end{tabular}
	\end{center}
	\label{tab: Tab3}
	\vspace{-3mm}
\end{table*}

\begin{figure*}[t]
	\begin{center}
		\includegraphics[width=1\linewidth]{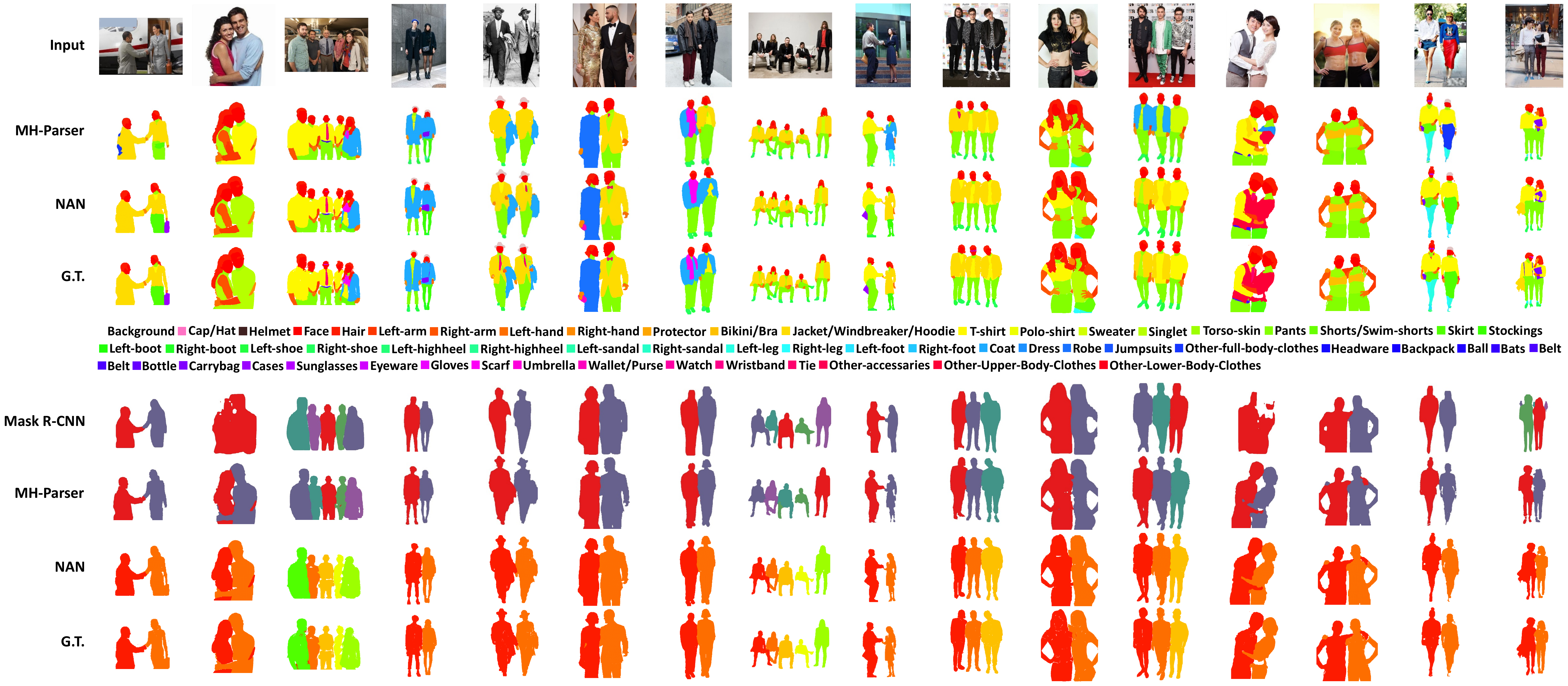}
	\end{center}
	\vspace{-5mm}
	\small
	\caption{\small Multi-human parsing qualitative comparison on the MHP v2.0 dataset. Best viewed in color.}
	\label{fig: Fig5_1}
	\vspace{-2mm}
\end{figure*}

The performance comparison of the proposed NAN with two state-of-the-art methods in terms of $\mathrm{AP}^{p}$@IoU=0.5, $\mathrm{AP}^{p}_{vol}$ and PCP@IoU=0.5 on the MHP v2.0 testing set is reported in Tab.~\ref{tab: Tab3}. Following~\cite{li2017towards}, we conduct experiments under three settings: \textbf{All} reports the evaluation over the whole testing set; \textbf{Inter$_{20\%}$} reports the evaluation over the sub-set containing the images with top 20\% interaction intensity\footnote{For each testing image, we calculate the pair-wise instance bounding box IoU and use the mean value as the interaction intensity for each image.}; \textbf{Inter$_{10\%}$} reports the evaluation over the sub-set containing the images with top 10\% interaction intensity. Our NAN is significantly superior over other state-of-the-arts on setting-1. In particular, NAN improves the $2^{nd}$-best by $7.15\%$, $5.70\%$ and $5.27\%$ in terms of all metrics. For the more challenging scenarios with intensive interactions (setting-2, 3), NAN also consistently achieves the best performance. In particular, for \textbf{Inter$_{20\%}$} and \textbf{Inter$_{10\%}$}, NAN improves the $2^{nd}$-best by $5.23\%$, $4.65\%$ and $5.62\%$; $1.63\%$, $3.20\%$ and $3.33\%$ in terms of all metrics. This verifies the effectiveness of our NAN for multi-human parsing and understanding humans in crowded scenes. Moreover, NAN can rapidly process one 512$\times$512 image in about 1 second with acceptable resource consumption, which is attractive to real applications. This compares much favorably to MH-Parser~\cite{li2017towards} (14.94 img/s), which relies on separate and complex post-processing (including CRF~\cite{NIPS2011_4296}) steps.

\subsubsection{Qualitative Comparison}

\begin{figure}[t]
	\begin{center}
		\includegraphics[width=1\linewidth]{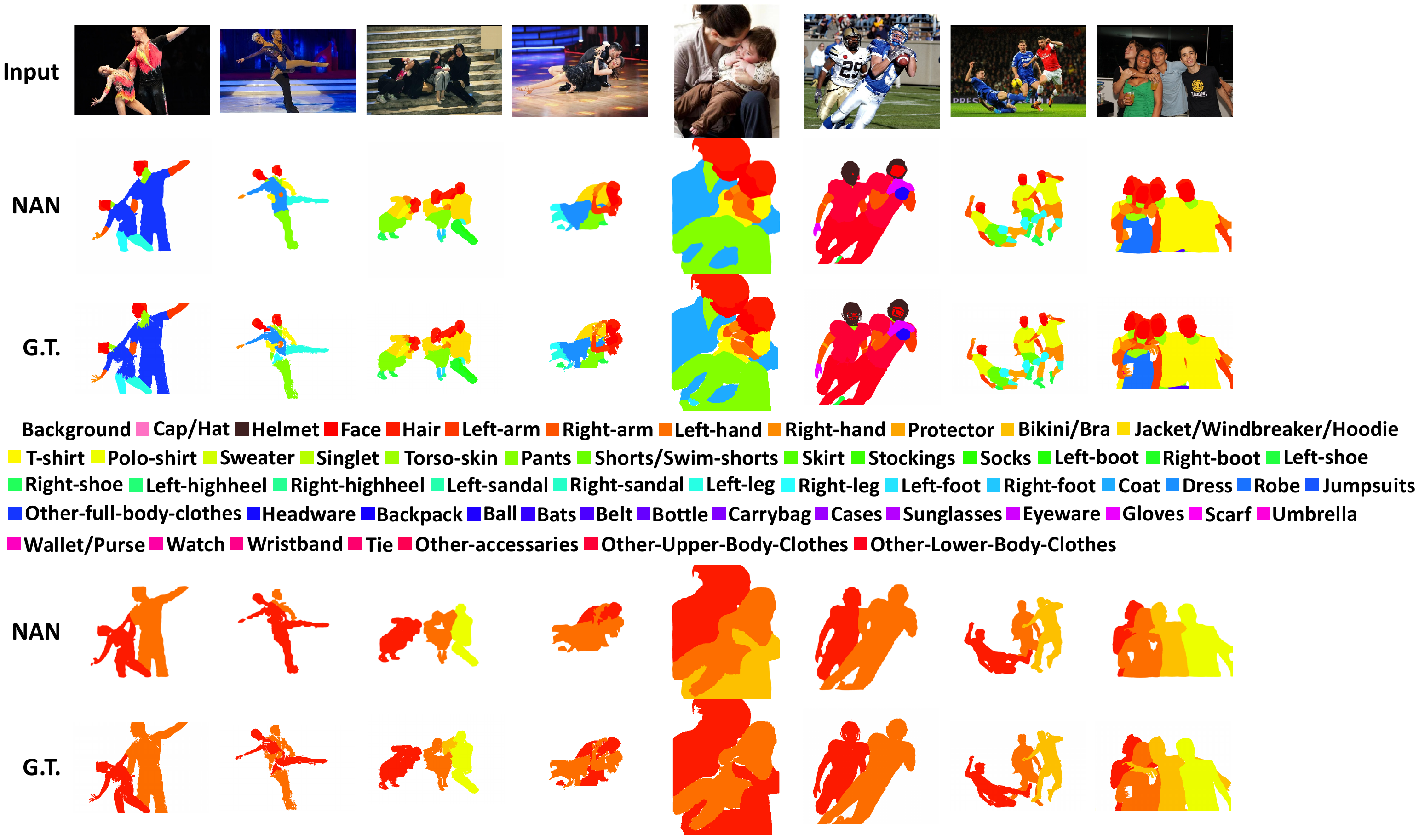}
	\end{center}
	\vspace{-4mm}
	\small
	\caption{\small Failure cases of multi-human parsing results by our NAN on the proposed MHP v2.0 dataset. Best viewed in color.}
	\label{fig: Fig5_2}
	\vspace{-3mm}
\end{figure}

Fig.~\ref{fig: Fig5_1} visualizes the qualitative comparison of the proposed NAN with two state-of-the-art methods and corresponding ground truths on the MHP v2.0 dataset. Note that Mask R-CNN~\cite{he2017mask} only offers silhouettes of different person instances, we only compare our instance-aware clustering results with it while comparing our holistic results with MH-Parser~\cite{li2017towards}. It can be observed that the proposed NAN performs well in multi-human parsing with a wide range of viewpoints, poses, occlusion, interactions and background complexity. The instance-agnostic parsing and instance-aware clustering predictions of NAN present high consistency with corresponding ground truths, thanks to the novel network structure and effective training strategy. In contrast, Mask R-CNN~\cite{he2017mask} suffers difficulties to differentiate entangled humans, while MH-Parser~\cite{li2017towards} struggles to generate fine-grained parsing results and clearly segmented instance masks. This further desmonstrates the effectiveness of the proposed NAN. We also show some failure cases of our NAN in Fig.~\ref{fig: Fig5_2}. As can be observed, humans in crowded scenes with heavy occlusion, extreme poses and intensive interactions are difficult to identify and segment. Some small-scale semantic categories within person instances are difficult to parse. This confirms that MHP v2.0 aligns with real-world situations and deserves more furture attention and research efforts.

\subsection{Evaluations on the MHP v1.0 Benchmark}

\begin{table}[t]
	\newcommand{\tabincell}[2]{\begin{tabular}{@{}#1@{}}#2\end{tabular}}
	\scriptsize
	\caption{\small Multi-human parsing quantitative comparison on the MHP v1.0~\cite{li2017towards} testing set.}
	\vspace{-4mm}
	\begin{center}
		\begin{tabular}{cccc}
			\hline
			\tabincell{c}{Method} & \tabincell{c}{$\mathrm{AP}^{p}_{0.5} (\%)$} & \tabincell{c}{$\mathrm{AP}^{p}_{vol} (\%)$} & \tabincell{c}{PCP$_{0.5} (\%)$} \\
			\hline
			DL~\cite{de2017semantic} & 47.76 & 47.73 & 49.21 \\
			MH-Parser~\cite{li2017towards} & 50.10 & 48.96 & 50.70 \\
			Mask R-CNN~\cite{he2017mask} & 52.68 & 49.81 & 51.87 \\
			\hline
			NAN & \textbf{57.09} & \textbf{56.76} & \textbf{59.91} \\
			\hline
		\end{tabular}
	\end{center}
	\label{tab: Tab4}
\end{table}

\begin{figure}[t]
	\begin{center}
		\includegraphics[width=1\linewidth]{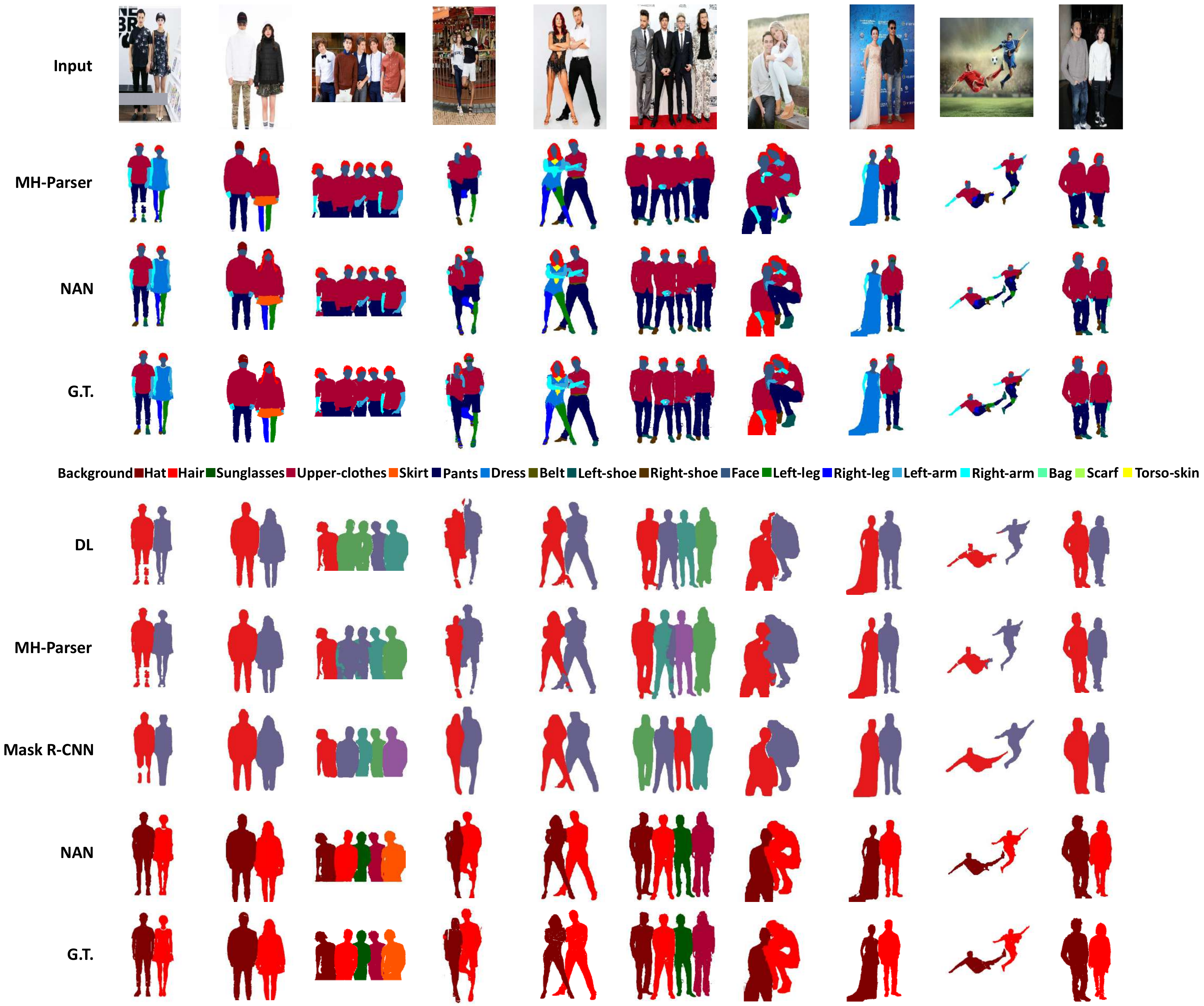}
	\end{center}
	\vspace{-4mm}
	\small
	\caption{\small Multi-human parsing qualitative comparison on the MHP v1.0~\cite{li2017towards} dataset. Best viewed in color.}
	\label{fig: Fig6}
\end{figure}

The MHP v1.0\footnote{The dataset is available at \url{http://lv-mhp.github.io/}} dataset is the first multi-human parsing benchmark, originally proposed by Li \emph{et al.}~\cite{li2017towards}, which contains 4{,}980 images annotated with 18 semantic labels. Annotation examples are visualized in Fig.~\ref{fig: Fig2} (b). The data are randomly organized into 3 splits, consisting of 3{,}000 training, 1{,}000 validation and 1{,}000 testing images with publicly available annotations. Evaluation systems report the $\mathrm{AP}^{p}$ and PCP over the testing set. Refer to~\cite{li2017towards} for more details.

The performance comparison of the proposed NAN with three state-of-the-art methods in terms of $\mathrm{AP}^{p}$@IoU=0.5, $\mathrm{AP}^{p}_{vol}$ and PCP@IoU=0.5 on the MHP v1.0~\cite{li2017towards} testing set is reported in Tab.~\ref{tab: Tab4}. With the nested adversarial learning of semantic saliency prediction, instance-agnostic parsing and instance-aware clustering, our method outperforms the $2^{nd}$-best by $4.41\%$ for $\mathrm{AP}^{p}_{0.5}$, $6.95\%$ for $\mathrm{AP}^{p}_{vol}$ and $8.04\%$ for PCP$_{0.5}$. Visual comparison of multi-human parsing results by NAN and three state-of-the-art methods is provided in Fig.~\ref{fig: Fig6}, which further validates the advantages of our NAN over existing solutions.

\subsection{Evaluations on the PASCAL-Person-Part Benchmark}

\begin{table}[t]
	\newcommand{\tabincell}[2]{\begin{tabular}{@{}#1@{}}#2\end{tabular}}
	\scriptsize
	\caption{\small Multi-human parsing quantitative comparison on the PASCAL-Person-Part~\cite{chen2014detect} testing set.}
	\vspace{-4mm}
	\begin{center}
		\begin{tabular}{ccccc}
			\hline
			\tabincell{c}{Method} & \tabincell{c}{$\mathrm{AP}^{r}_{0.5} (\%)$} & \tabincell{c}{$\mathrm{AP}^{r}_{0.6} (\%)$} & \tabincell{c}{$\mathrm{AP}^{r}_{0.7} (\%)$} & \tabincell{c}{$\mathrm{AP}^{r}_{vol} (\%)$} \\
			\hline
			MNC~\cite{dai2016instance} & 38.80 & 28.10 & 19.30 & 36.70 \\
			Li \emph{et al.}~\cite{li2017holistic} & 40.60 & 30.40 & 19.10 & 38.40 \\
			\hline
			NAN & \textbf{59.70} & \textbf{51.40} & \textbf{38.00} & \textbf{52.20} \\
			\hline
		\end{tabular}
	\end{center}
	\label{tab: Tab5}
\end{table}

\begin{figure}[t]
	\begin{center}
		\includegraphics[width=1\linewidth]{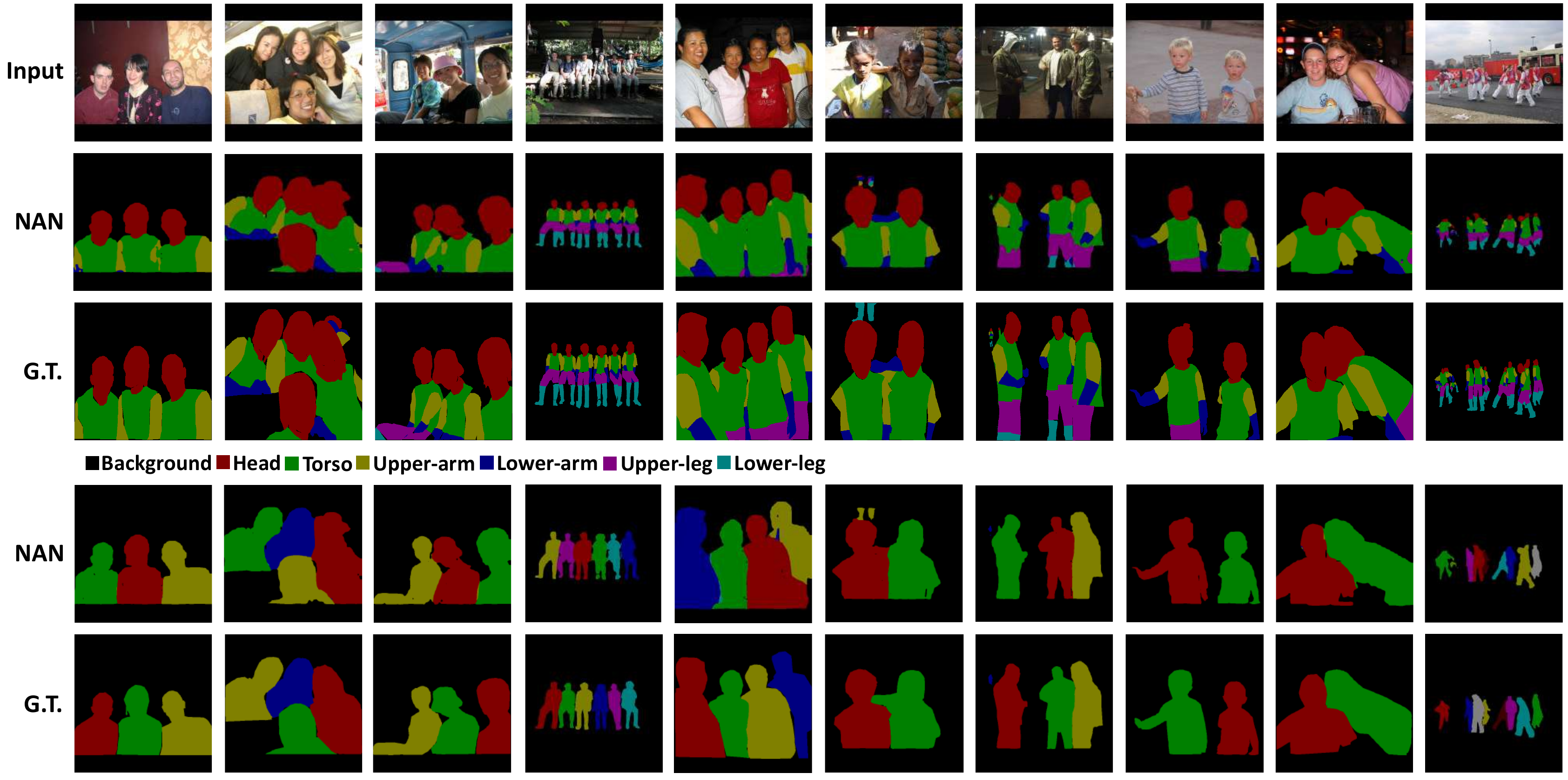}
	\end{center}
	\vspace{-4mm}
	\small
	\caption{\small Multi-human parsing qualitative comparison on the PASCAL-Person-Part~\cite{chen2014detect} dataset. Best viewed in color.}
	\label{fig: Fig7}
\end{figure}

The PASCAL-Person-Part\footnote{The dataset is available at \url{http://www.stat.ucla.edu/~xianjie.chen/pascal\_part\_dataset/pascal\_part.html}}~\cite{chen2014detect} dataset is a set of additional annotations for PASCAL-VOC-2010~\cite{everingham2010pascal}. It goes beyond the original PASCAL object detection task by providing pixel-wise labels for six human body parts, \emph{i.e.}, \emph{head}, \emph{torso}, \emph{upper-/lower-arms}, and \emph{upper-/lower-legs}. The rest of each image is considered as background. There are 3{,}535 images in the PASCAL-Person-Part~\cite{chen2014detect} dataset, which is split into separate training set containing 1{,}717 images and testing set containing 1{,}818 images. For fair comparison, we report the $\mathrm{AP}^{r}$ over the testing set for multi-human parsing. Refer to~\cite{chen2016attention, xia2016zoom} for more details. 

The performance comparison of the proposed NAN with two state-of-the-art methods in terms of $\mathrm{AP}^{r}$@IoU=$k_{k=_{0.5}^{0.7}}$ and $\mathrm{AP}^{r}_{vol}$ on the PASCAL-Person-Part~\cite{chen2014detect} testing set is reported in Tab.~\ref{tab: Tab5}. Our method dramatically surpasses the $2^{nd}$-best by $18.90\%$ for $\mathrm{AP}^{r}_{0.7}$ and $13.80\%$ for $\mathrm{AP}^{r}_{vol}$. Qualitative multi-human parsing results by NAN are visualized in Fig.~\ref{fig: Fig7}, which possess a high concordance with corresponding ground truths. This again verifies the effectiveness of our method for human-centric analysis.

\subsection{Evaluations on the Buffy Benchmark}

\begin{table}[t]
	\newcommand{\tabincell}[2]{\begin{tabular}{@{}#1@{}}#2\end{tabular}}
	\scriptsize
	\caption{\small Instance segmentation quantitative comparison on the Buffy~\cite{vineet2011human} dataset episode 4, 5 and 6.}
	\vspace{-4mm}
	\begin{center}
		\begin{tabular}{cccc}
			\hline
			\tabincell{c}{Method} & \tabincell{c}{F (\%)} & \tabincell{c}{B (\%)} & \tabincell{c}{Ave. (\%)} \\
			\hline
			Vineet \emph{et al.}~\cite{vineet2011human} &  - & - & 62.40 \\
			Jiang \emph{et al.}~\cite{jiang2016detangling} & 68.22 & 69.66 & 68.94 \\
			MH-Parser~\cite{li2017towards} & 71.11 & 71.94 & 71.53 \\
			\hline
			NAN & \textbf{77.24} & \textbf{79.92} & \textbf{78.58} \\
			\hline
		\end{tabular}
	\end{center}
	\label{tab: Tab6}
\end{table}

\begin{figure}[t]
	\begin{center}
		\includegraphics[width=1\linewidth]{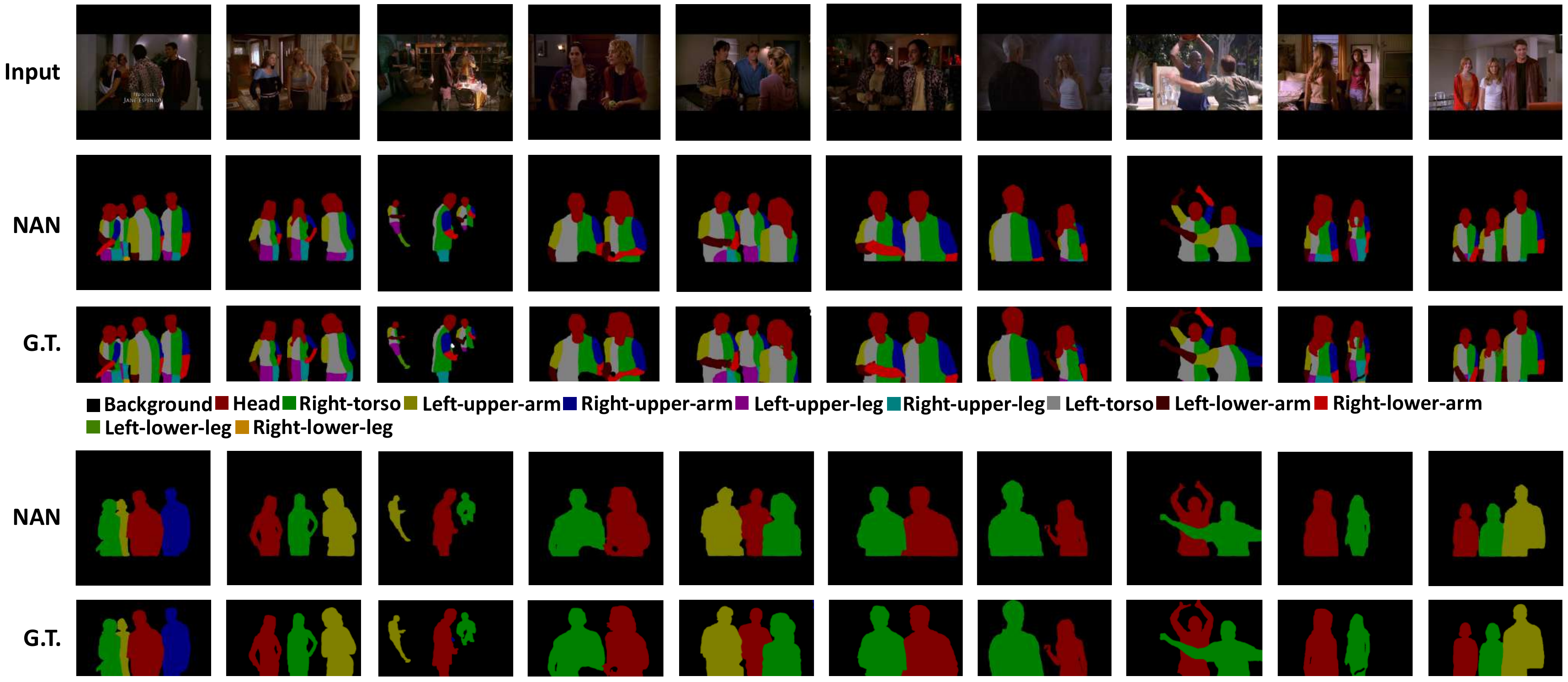}
	\end{center}
	\vspace{-4mm}
	\small
	\caption{\small Qualitative instance-agnostic parsing (upper panel) and instance-aware clustering (lower panel) results by NAN on the Buffy~\cite{vineet2011human} dataset. Best viewed in color.}
	\label{fig: Fig8}
\end{figure}

The Buffy\footnote{The dataset is available at \url{https://www.inf.ethz.ch/personal/ladickyl/Buffy.zip}}~\cite{vineet2011human} dataset was released in 2011 for human parsing and instance segmentation, which contains 748 images annotated with 12 semantic labels. The data are randomly organized into 2 splits, consisting of 452 training and 296 testing images with publicly available annotations. For fair comparison, we report the \textbf{F}orward (F) and \textbf{B}ackward (B) scores~\cite{jiang2016detangling} over the episode 4, 5 and 6 for instance segmentation evaluation. Refer to~\cite{vineet2011human, jiang2016detangling} for more details.

The performance comparison of the proposed NAN with three state-of-the-art methods in terms of F and B scores on the Buffy~\cite{vineet2011human} dataset episode 4, 5 and 6 is reported in Tab.~\ref{tab: Tab6}. Our NAN consistently achieves the best performance for all metrics. In particualr, NAN significantly improves the $2^{nd}$-best by $6.13\%$ for F score and $7.98\%$ for B score, with an average boost of $7.05\%$. Qualitative instance-agnostic parsing and instance-aware clustering results by NAN are visualized in Fig.~\ref{fig: Fig8}, which well shows the promising potential of our method for fine-grained understanding humans in crowded scenes.

\section{Conclusions}

In this work, we presented ``\textbf{M}ulti-\textbf{H}uman \textbf{P}arsing (MHP v2.0)", a large-scale multi-human parsing dataset and a carefully designed benchmark to spark progress in understanding humans in crowded scenes. MHP v2.0 contains 25{,}403 images, which are richly labelled with 59 semantic categories. We also proposed a novel deep \textbf{N}ested \textbf{A}dversarial \textbf{N}etwork (NAN) model to address this challenging problem and performed detailed evaluations of the proposed method with current state-of-the-arts on MHP v2.0 and several other datasets. We envision the proposed MHP v2.0 dataset and the baseline method would drive the human parsing research towards real-world application scenario with simultaneous presence of multiple persons and complex interactions among them. In future, we will continue to take efforts to construct a more comprehensive multi-human parsing benchmark dataset with more images and more detailed semantic category annotations to further push the frontiers of multi-human parsing research.

\vspace{-2mm}
\section*{Acknowledgement}
\vspace{-1mm}

The work of Jian Zhao was partially supported by \textbf{C}hina \textbf{S}cholarship \textbf{C}ouncil (CSC) grant 201503170248.

The work of Jiashi Feng was partially supported by NUS startup R-263-000-C08-133, MOE Tier-I R-263-000-C21-112, NUS IDS R-263-000-C67-646 and ECRA R-263-000-C87-133.

{\small
\bibliographystyle{ieee}
\bibliography{egbib}
}

\end{document}